\documentclass[twocolumn,prx,superscriptaddress]{revtex4-1}
\usepackage[utf8]{inputenc}
\usepackage{xcolor}
\usepackage{bm}
\usepackage{graphicx}
\usepackage{physics}
\usepackage{amsmath}
\usepackage{amssymb}
\usepackage{times}
\usepackage{enumitem}
\usepackage{array}
\usepackage{dcolumn}
\usepackage{comment}
\usepackage{tikz}
\usetikzlibrary{arrows.meta,positioning}
\usepackage{adjustbox}

\usepackage[colorlinks=true,citecolor=blue]{hyperref}

\begin{document}

\title{All You Need Is Non-Commutative Words}

\author{Carla M. Quispe Flores}
\email{carlaquispeflores@mines.edu}
\affiliation{Department of Physics, Colorado School of Mines, Golden, Colorado 80401, USA}

\author{Stanley Salvatierra}
\thanks{C. M. Quispe Flores and S. Salvatierra contributed equally to this work.}
\affiliation{Lookia MX, OR 0401, BO}

\author{Renan Cabrera}
\affiliation{Yardley, PA 19067, USA}

\begin{abstract}
We represent lexical tokens as unitary matrices and encode each sentence as their ordered product. The noncommutativity of matrix product captures word order without positional encodings (PEs). The same algebra yields several capabilities, including antisymmetric self-attention with no query, key, or value projections, and parallel composition of variable-length text chunks at a reduced attention cost. Furthermore, it provides a canonical-coset readout layer that  encodes all true unitary degrees of freedom compactly, while supporting continual learning through nested group extensions that enlarge the operator space with each new task preserving prior representations exactly. Across standard text-classification benchmarks, the method matches or exceeds bag-of-words baselines. Achieving higher accuracy on IMDB and comparable performance on AG News. Notably, this is accomplished by replacing the conventional $\sim$30,000-dimensional vocabulary space with a dense, 64-parameter real-valued encoding, highlighting the expressive efficiency of our parameterization. 

\end{abstract}

\maketitle

\newcommand{\avg}[1]{\langle #1 \rangle} %expectation value

\section{Introduction}
Although sentences are inherently structured by sequence, widely adopted text representations exhibit permutation invariance. Bag-of-words (BoW) \cite{weinberger2009feature} aggregate word vectors by addition or averaging, and are therefore invariant under any permutation of the input. Transformer-based large language models (LLMs) encounter a complementary problem: the self-attention mechanism is insensitive to token ordering. Thus, positional information must be added separately, through additive encodings, relative-position biases, or rotary embeddings (RoPE) \cite{su2024roformer}. In both cases sequential structure is grafted onto the model via added parameters.

In contrast, group theory offers a natural framework to represent sequence structure intrinsically. Unitary evolution recurrent networks~\cite{arjovsky2016unitary} and their efficient parametrizations~\cite{jing2017tunable} constrained hidden-to-hidden weights to the unitary group, resolving the vanishing- and exploding-gradient problem \cite{pascanu2013difficulty}; non-normal extensions~\cite{kerg2019non} relaxed strict orthogonality to admit transient dynamics and reached state-of-the-art character-level modeling among non-gated architectures. In all of these models, the unitary matrix is a hidden-state transition operator: order is handled by the recurrence, and the word embedding remains a conventional vector.

Non-commutativity has also been recognized as a mechanism for encoding order directly. PaTH~\cite{yang2026path} represents position through accumulated, input-dependent Householder reflections, and such non-commutative products substantially improve state tracking and length extrapolation; yet the products serve strictly as a position code inside standard attention, with query, key, and value projections intact. Closer to our view, Bernardy and Lappin~\cite{bernardy2022unitary,bernardy2022assessing} encode words as orthogonal matrices obtained by exponentiating skew-symmetric generators and compose sentences by sequential matrix products, achieving rule-like accuracy on subject-verb agreement and generalized Dyck bracket matching; their model, however, is a purely recurrent cell over real orthogonal matrices, evaluated only on syntactic prediction. The appeal of algebraically structured sentence spaces extends beyond order: \cite{zhang2024learning} obtain interpretable and controllable sentence representations by disentangling the latent space of a language autoencoder with invertible neural networks, though the geometry there is learned from data, not fixed by a group.

Quantum-inspired language models pursue a complementary direction, modeling sentence semantics as the sequential unitary evolution of lexical superposition states and reporting parameter-efficient gains on question answering and text classification~\cite{fan2024quantum,nausheen2025quantum}. In contrast, Anschuetz \emph{et al.}~\cite{anschuetz2023interpretable} study genuinely quantum models: recurrent architectures built from unitary dynamics with non-Gaussian measurements, and proved an unconditional memory separation over trainable classical sequence models. Identifying quantum contextuality as the interpretable source of the advantage. Most of these efforts remain small in scale~\cite{nausheen2025quantum}, and they retain the quantum state as the object of representation: the sentence is a density matrix updated word by word through a serial recursion that admits no parallel regrouping.

\begin{figure*}[t]%[htbp]
\centering
\includegraphics[scale=0.9,angle=0]{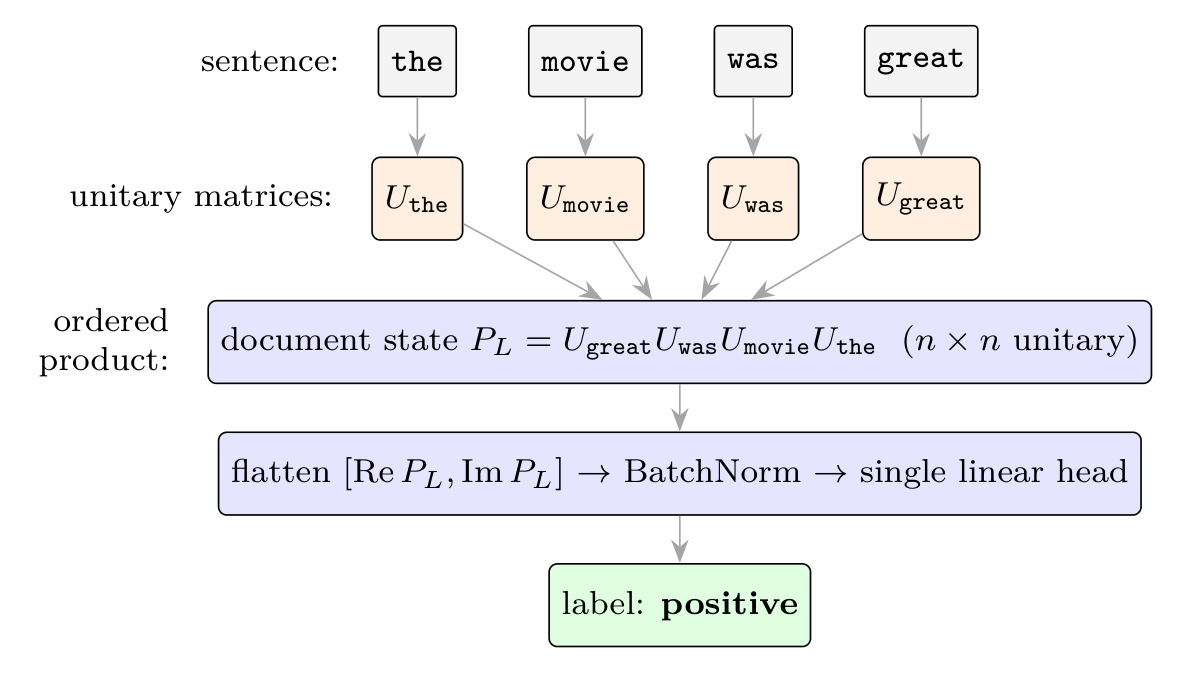}
\caption{Word order is captured natively by representing each word as a unitary matrix and composing the sentence via their ordered product. Because matrix multiplication is noncommutative, the resulting document state $P_L$ preserves token ordering without positional encodings (PEs). A flattened readout followed by a single linear head then maps this state to the target label.}
\label{fig:main-model}
\end{figure*}

Despite these effords, no existing framework simultaneously represents every token as a learnable unitary group element, encodes word order through matrix non-commutativity instead of added positional signals, and derives self-attention, chunk-parallel composition, and continual learning as direct consequences of the same algebraic structure. In this work we introduce such an architecture, implemented as a single forward network. Input token identifiers are mapped to real coordinates in a Chevalley basis; the coordinates assemble one Hermitian generator per word; matrix exponentiation, regulated by an adaptive per-word rotation budget $\varepsilon_w$, turns each generator into a unitary word operator; a time-ordered matrix product composes the operators into a single document unitary; and a canonical-coset readout extracts its intrinsic coordinates, which a batch-normalized linear head classifies, as shown in Fig.~\ref{fig:main-model}.

Order sensitivity follows from the non-commutativity of the composed operators, requiring neither positional encodings nor causal masks. The same algebra yields an exactly antisymmetric self-attention score computed from one learned Hermitian matrix, with no query, key, or value projections; group closure lets variable-length chunks be composed in parallel at reduced attention cost; and the nesting $U(n) \subset U(n+k)$ supports continual learning that preserves prior tasks exactly. This construction is inspired by the mathematics of quantum mechanics, but every computation here is classical and we claim no quantum advantage. The model matches bag-of-words baselines on standard text-classification benchmarks, while replacing a $\sim$30{,}000-dimensional vocabulary space with a dense 64-coordinate encoding.

% NOTE (Carla): please add \label{fig:main-model} to the main_model figure environment; the current draft references an empty Fig.().

The remainder of this paper is organized as follows. Section~\ref{sec:methods} develops the framework and the full network architecture, Section~\ref{sec:results} presents and discusses the results, and Section~\ref{sec:conclusion} concludes with contributions and outlook.

\section{Methods}\label{sec:methods}

\subsection{Input Representation}
The framework maps discrete vocabulary words to a set of unitary operators with continuous parametrization. Each vocabulary word is assigned a learned Hermitian generator $H_w \in \mathbb{C}^{n \times n}$,  which is then exponentiated to produce a unitary matrix $U_w$ with a per-word rotation budget $\varepsilon_w$
\begin{align}
\label{eq-Uw}
U_w = \exp(i \varepsilon_w \hat{H}_w) \in \mathrm{U}(n), \qquad \hat{H}_w = \hat{H}_w^\dagger. 
\end{align}
We normalize $\widehat H_w$ to unit Frobenius norm, $\lVert\widehat H_w\rVert_F=1$. The effective generator $\varepsilon\widehat H_w$ has
eigenvalues $\lambda_a$ satisfying $\sum_a \lambda_a^2=\varepsilon_w^2$, hence $|\lambda_a|\le\varepsilon_w$. Each word induces a small rotation, with the per-word rotation budget $\varepsilon_w$ scaling the rotation magnitude. A detailed derivation of the corresponding $\varepsilon_w$ bound is provided in Appendix~\ref{app:bounding-eigenphase}.

The Hermitian generators are defined via a Chevalley basis~\cite{cahn2014semi}, which provides a complete, real, orthonormal coordinate system for the Lie algebra. Any real linear combination of its elements is therefore a valid Hermitian operator. In this work, we consider the Lie algebra $\mathfrak{u}(n)$ with $n = 8$, whose Chevalley basis comprises 64 Hermitian generators $\{T_k\}$: 8 diagonal $E_{ii}$, 28 symmetric $X_{ij} = (E_{ij} + E_{ji})/\sqrt{2}$, and 28 antisymmetric $Y_{ij} = \mathrm{i}(E_{ij} - E_{ji})/\sqrt{2}$ for $i < j$,
\begin{align}
\{T_k\}_{k=1}^{64} = \{E_{ii}\}_{i=1}^8 \cup \{X_{ij}\}_{i<j} \cup \{Y_{ij}\}_{i<j}.
\end{align}
These 64 basis elements ${T_k}$ are $8 \times 8$ Hermitian matrices spanning the 64-dimensional real vector space of all $8 \times 8$ Hermitian matrices, so any Hermitian generator $H_w$ can be written uniquely as a real linear combination of them
\begin{align}
H_w = \sum_{k=1}^{64} (C_w)_k T_k,
\end{align}
where $C_w \in \mathbb{R}^{64}$ is the coordinate vector for word $w$ in the Chevalley basis. $C_w$ can be supplied in two ways: the \emph{free-table} method, a free lookup table learns unique parameters per word, acting as an independent Hermitian table that costs $1{,}279{,}872$ parameters across the vocabulary. Alternatively, in the \emph{distilled-map} method, a shared linear map projects a frozen, 300-dimensional Word2Vec embedding $v_w$ (GoogleNews) into the same 64-dimensional coordinate space spanned by the Chevalley basis
\begin{equation}
C_w \;=\; W\,v_w + b
\label{eq:distill-map}
\end{equation}
with $W\in\mathbb{R}^{64\times300}$ and $b\in\mathbb{R}^{64}$ ($19{,}328$ shared
parameters in total), trained end-to-end on the classification loss, and therefore independent of vocabulary size. Thus, ``distillation'' in this context refers to reusing pretrained features via a learned projection into the operator algebra.

%%%%%%%%%%%%%%%%%%%%%%%%%%%%%%%%%%%%%%%%%%%%
\subsection{Adaptive Per-Word Rotation Budget}
A global rotation budget $\varepsilon$ treats all words equally, but content words (``excellent'', ``terrible'') carry strong discriminative signals and therefore benefit from larger rotations, whereas function words (``the'',
``and'') contribute little and should rotate less. Therefore, we let each word predict its own rotation budget from its embedding as
\begin{align}
\label{eq-adaptive-eps}
  \varepsilon_w = \varepsilon \cdot \mathrm{softplus}(u \cdot v_w + c),
\end{align}
this introduces only $\sim$$301$ extra parameters: $u \in \mathbb{R}^{300}$ is a learned weight vector, $c \in \mathbb{R}$ is a learned scalar bias, $v_w \in \mathbb{R}^{300}$ is the frozen Word2Vec \cite{mikolov2013efficient} embedding for word $w$, and $\text{softplus}(x) = \log(1 + e^x)$ ensures positivity. The accumulated eigenphase for a sequence of lenght $L$ accumulates as $\sum_{i=1}^L \varepsilon_w \lambda_a$, where $\lambda_a$ are the eigenvalues of the normalized generator. This linear accumulation requires careful control of $\varepsilon_w$ to prevent phase aliasing at long sequences. A fixed global $\varepsilon$ induces systematic over-rotation in long documents, driving phases toward wrap-around, whereas the learned $\epsilon_w$ dynamically allocates phase budget magnitude based on word importance.
\subsection{Order for Free: Non-Commutative Matrix Product}

This stage composes word unitaries $U_w$ into a document representation $P_L$ by
time-ordered matrix product. Group closure guarantees that the whole document collapses to a single unitary $P_L\in \mathrm{U}(8)$, so variable-length sequences are supported with no padding to a fixed dimension. 

A sequence of $j$ words $\{w_1, w_2, \ldots, w_j\}$ is represented by the prefix products
\begin{align}
\label{eq-prefix}
P_j = U_{j} U_{j-1} \cdots U_{1} \in \mathrm{U}(8), \quad i = 1, 2, \ldots, j,
\end{align}
which serve as the ``hidden states'' of the construction of the document operator $P_L$. Because matrix multiplication generally does not commute ($U_i U_j \neq U_j U_i$), the product $P_L$ depends on word order. Swapping words changes the final unitary, making the representation order-sensitive without requiring positional encodings. Then the final document operator $P_L$ is the
single object flattened into
$[\Re P_L,\Im P_L]\in\mathbb{R}^{128}$.

\subsection{QKV-Free Self-Attention}
Conventional transformer attention computes a score for each pair of positions using the scaled dot product between a query and a key projections \cite{vaswani2017attention} $S_{ij}=q_i\cdot k_j/\sqrt{d_k}$, with $d_k$ denoting the dimensionality of the key (and query) vectors and serving as a normalization factor to stabilize gradients during training. Because the dot product is permutation-invariant, positional encodings must be added by hand, and capturing directed relations ($S_{ij}\neq S_{ji}$) requires three full projection matrices, roughly $3d_{\text{model}}^2$ parameters. Our operator representation removes the first overhead: the prefix product $P_i$ already encodes word order through the non-commutativity of matrix product, so attention runs directly over prefix products with no positional encoding.
For Hermitian $A = A^\dagger$ and unitary prefix operators $P_i$ and $P_j$, the self-attention score is exactly antisymmetric
\begin{align}
\label{Sij}
S_{ij} \;=\; \frac{1}{n}\, \mathrm{Im}\operatorname{Tr}\!\big(A\,P_i^\dagger P_j\big).
\end{align}
This yields a parameter-efficient, direction-sensitive scoring mechanism with $n^2 = 64$ parameters in a single Hermitian learned matrix $A$, rather than three projection matrices ($\sim 3 d_{\text{model}}^2$ in a standard Q/K/V stack). This is consistent with prior work showing that shared query--key maps are necessarily symmetric \cite{kayyam2026transformers} and that directional information resides in the antisymmetric part of attention \cite{saponati2025underlying}, here recovered from first principles.

\begin{figure*}[t]%[htbp]
\centering
\includegraphics[scale=0.56,angle=0]{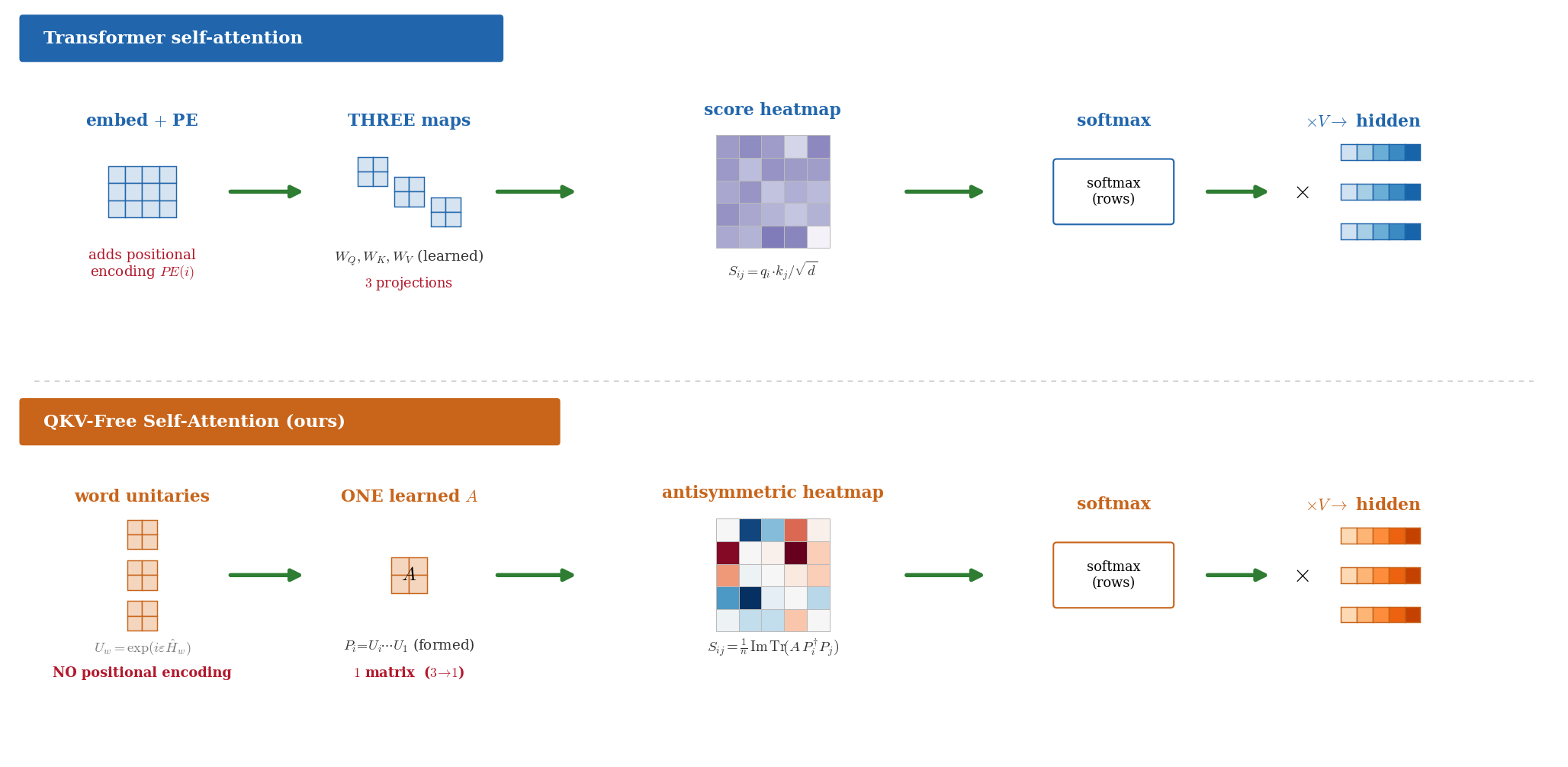}
\caption{Conventional self-attention mechanics (blue) versus our
QKV-Free Self-Attention (orange). The former needs positional encodings and the
projections $W_Q$, $W_K$, $W_V$ to score token pairs; the latter scores
unitary prefix products directly through one Hermitian matrix $A$, with
order supplied by non-commutativity and antisymmetry of $S_{ij}$ built in.}
\label{fig:attn-sidebyside}
\end{figure*}

Appendix~\ref{specA} gives the spectral interpretation of $A$ as a weighted sum of overlaps between reference states transported by $P_i$ and $P_j$, together with its density-operator interpretation. Figure~\ref{fig:attn-sidebyside} compares the two self-attention mechanisms step by step, highlighting the removal of positional encodings and the $W_Q/W_K/W_V$ projections, and the replacement of these components by a single learned Hermitian operator $A$.

\subsection{Scalability: Variable-Length Document Chunking}
Group closure offers a path to long-context scalability, an objective shared by linear attention \cite{katharopoulos2020transformers} and structured state-space models \cite{gu2023mamba,gu2021efficiently}. In our setting, we let $c$ denote the number of consecutive words per chunk, partitioning a document of length $L$ into $L/c$ chunks. Since the ordered product of $c$ consecutive $n \times n$ unitaries remains a single unitary, a chunk collapses into
\begin{align}
\label{eq-chunk}
\widetilde P_k = U_{ck} U_{ck-1} \cdots U_{c(k-1)+2} U_{c(k-1)+1}.
\end{align}
Here $\widetilde P_k$ is an independent ordered product with no state propagated between chunks. As a result, matrix associativity allows these ordered products to be evaluated in parallel at any granularity, while group closure ensures the resulting document remains a valid unitary operator. However, $\widetilde P_k$ accumulates rotation budget across its $c$ words, much like the final readout $P_L$ accumulates across all $L$. To ensure chunking remains wrap-safe, we set the within-chunk budget to $\varepsilon_w = \varepsilon/c$. This choice systematically bounds the total rotation budget per ordered product: by dividing the primary rotation budget $\varepsilon$ by the chunk size $c$, we guarantee that the total accumulated phase across the entire chunk cannot exceed the target maximum $c\cdot(\varepsilon/c) = \varepsilon$, safely preventing the phase from wrapping around. This requires that $\varepsilon$ sit below the wrap threshold.

The chunking likewise reduces the attention cost: the self-attention score $S_{ij}$ operates over the $L/c$ ordered products, reducing the score-pair computation cost by a factor of $1/c^2$ relative to the unchunked ($c=1$) baseline. Consequently, $c$ governs the trade-off between attention resolution and computational efficiency. While this indicates a pathway to sub-quadratic token-level attention, our claims remain limited to the empirical score-pair counts.

\subsection{Canonical-Coset Readout Layer}
\label{sec:coset-readout}

The standard readout layer flattens the document operator $P_L \in U(n)$ into its real and imaginary parts
\begin{align}
\label{eq-u_flat}
u_{\text{flat}} \;=\; [\,\mathrm{Re}\,P_L,\; \mathrm{Im}\,P_L\,] \;\in\; \mathbb{R}^{2n^2}
\end{align}
which contains $2n^2$ real numbers. For $n = 8$, Eq.(\ref{eq-u_flat}) yields a 128-dimensional vector. However, $P_L$ has only $n^2 = 64$ real degrees of freedom, with the unitarity constraint $P_L^\dagger P_L = I$ tying exactly half of the entries. The 128-flatten is therefore an overcomplete chart, and the linear head must implicitly learn to navigate the constrain.

The canonical coset decomposition~\cite{gilmore2011group,cabrera2010canonical} provides the $n^2$ intrinsic coset coordinates. The idea is to record $P_L$ column by column, retaining only the genuinely new information each column carries. Every element of $U(n)$ factors as
\begin{small}
\begin{multline}
\label{eq-chain}
U(n) = \frac{U(n)}{U(n-1)\otimes U(1)}\;\frac{U(n-1)}{U(n-2)\otimes U(1)} \\
\cdots\,\frac{U(2)}{U(1)\otimes U(1)}\; U(1)^{\otimes n}
\end{multline}
\end{small}
where $U(1)^{\otimes n}$ is a diagonal matrix of $n$ residual phases; the $k$-th quotient captures the freedom remaining in the $k$-th column once all previous columns are fixed, and is parametrized by a single complex vector $\lvert X_k \rangle \in \mathbb{C}^{n-k}$ 
{\small
\begin{multline}
\label{eq-U_cotient}
\frac{U(n-k+1)}{U(n-k)\otimes U(1)} = I_{k-1} \,\oplus\, \\
\begin{pmatrix}
\sqrt{1 - \langle X_k \vert X_k \rangle} & -\langle X_k \rvert \\[4pt]
\lvert X_k \rangle & \sqrt{\,I_{n-k} - \lvert X_k \rangle\langle X_k \rvert\,}
\end{pmatrix}
\end{multline}
}
Two aspects of Eq.~(\ref{eq-U_cotient}) capture the essential structure. First, the top-left entry enforces a ball constraint: $\sqrt{1 - \langle X_k \vert X_k \rangle}$ is real only if $\langle X_k \vert X_k \rangle \leq 1$, so the $2(n-k)$ real Cartesian components of $\lvert X_k \rangle$ range over the closed unit ball $B^{2(n-k)}$. Second, the coordinates carry direct geometric meaning: $\lvert X_k \rangle$ is the part of the $k$-th column that tilts away from its target axis $\lvert e_k \rangle$, so the radius $r_k = \sqrt{\langle X_k \vert X_k \rangle}$ runs from $r_k = 0$ (column aligned) to $r_k = 1$ (column fully perpendicular), with $\sqrt{1 - r_k^2}$ the surviving alignment. In practice each factor is computed as a product of two Householder reflections~\cite{cabrera2010canonical}. The construction and the one-line derivation of the ball constraint from column unit-length are given in Appendix~\ref{app:coset-construction}.

Counting the coordinates confirms the exact intrinsic dimension of $U(n)$ is complete: $\sum_{k=1}^{n-1} 2(n-k) \;+\; n \;=\; n(n-1) + n \;=\; n^2$. For $n = 8$, the shells $B^{14}, B^{12}, B^{10}, B^{8}, B^{6}, B^{4}$ and $B^{2}$, contribute $14 + 12 + 10 + 8 + 6 + 4 + 2 = 56$ ball coordinates and the diagonal phases the remaining $8$. Hence, the readout vector $u_{\text{coset}} \in \mathbb{R}^{64}$ is batch-normalized \cite{ioffe2015batch} and passed to the linear head.
%%%%%%%%%%%%%%%%%%%%%%%%%%%%%%%%%%%%%%%%%%%%%%%%%%%%%%%%%%%%%%%%%%%%%%%%%
\subsection{Coset-Tower and Continual Learning}
\label{sec:coset-tower}
Continual-learning methods are commonly grouped into regularization approaches, which penalize movement of weights important to prior tasks~\cite{kirkpatrick2017overcoming}; replay approaches, which retain or regenerate past examples~\cite{lopez2017gradient}; and parameter-isolation approaches, which allocate disjoint capacity per task~\cite{rusu2016progressive,mallya2018packnet,serra2018overcoming}. The coset tower belongs to the third family, but the isolation is exact and algebraic rather than heuristic.

The canonical coset decomposition of Section~\ref{sec:coset-readout} plays a second role in the architecture, beyond re-coordinatizing a fixed document operator $P_L$, it organizes the growth of the operator space. Unitary groups nest canonically, $U(n) \subset U(n+k)$, and the decomposition makes the nesting explicit: each step of the resulting \emph{coset tower} appends one shell of new coordinates while everything already trained persists intact as the top-left block. Extending $U(n)$ to $U(n+k)$ adds
\begin{align}
\label{eq-shell-dim}
(n+k)^2 - n^2 \;=\; 2nk + k^2
\end{align}
new real coordinates. The \emph{shell}, and a trained block $G_A \in U(n)$ embeds as
\begin{align}
\label{eq-embed}
G_A \;\longmapsto\;
\begin{pmatrix} G_A & 0 \\ 0 & I_k \end{pmatrix}
\;\in\; U(n+k).
\end{align}
The embedding~\eqref{eq-embed} is a group homomorphism, $\operatorname{diag}(G, I_k)\,\operatorname{diag}(G', I_k) = \operatorname{diag}(GG', I_k)$: products of embedded operators are embedded products. Consequently, every prefix product, $P_j$, and every document operator, $P_L$, assembled from embedded word operators is itself exactly of the form~\eqref{eq-embed}.
 
The coset factorization organizes the new coordinates, now applied block-wise rather than column by column: every $V \in U(n+k)$ factors as
\begin{align}
\label{eq-coset-factor}
V \;=\; \Omega(X)
\begin{pmatrix} G & 0 \\ 0 & G' \end{pmatrix},
\quad G \in U(n),\; G' \in U(k),
\end{align}
where the coset representative $\Omega(X) \in U(n+k)/\bigl(U(n)\otimes U(k)\bigr)$ carries the block form ~\cite{gilmore2011group}
\begin{align}
\label{eq-Omega}
\Omega(X) &\;=\;
\begin{pmatrix}
\sqrt{\,I_n - X X^{\dagger}\,} & X \\[2pt]
-\,X^{\dagger} & \sqrt{\,I_k - X^{\dagger} X\,}
\end{pmatrix},
\\
X &\in \mathbb{C}^{n\times k},
\qquad
X^{\dagger} X \preceq I_k.
\nonumber
\end{align}
The constraint $X^{\dagger}X \preceq I_k$ is a \emph{matrix ball}: the singular values of $X$ lie in $[0,1]$ and measure the mixing between the old and new sectors, exactly as the radii of the vector balls of Section~\ref{sec:coset-readout} measure the tilt of a single column; setting $k = 1$ in~\eqref{eq-Omega} recovers a factor of the form~\eqref{eq-U_cotient}. The count agrees with~\eqref{eq-shell-dim}: the coset contributes $\dim_{\mathbb{R}} U(n+k)/\bigl(U(n)\otimes U(k)\bigr) = 2nk$ real coordinates and the internal $U(k)$ block the remaining $k^2$; equivalently, iterating the elementary chain~\eqref{eq-chain} resolves the shell into $k$ vector shells plus $k$ phases (for $n = 8$, $k = 4$: $B^{22}$, $B^{20}$, $B^{18}$, $B^{16}$ and four phases, $76 + 4 = 80$). Two facts anchor the tower: $\Omega(0) = I_{n+k}$, so a vanishing shell ($X = 0$ and $G' = I_k$) returns $V$ exactly to the image of the embedding~\eqref{eq-embed}; and this configuration is reached by a linear projection of the shell coordinates, not by an optimization.
 
The Hermitian generators of the model decompose in the same blocks,
\begin{align}
\label{eq-H-blocks}
H \;=\;
\begin{pmatrix} H_A & M \\ M^{\dagger} & H_k \end{pmatrix},
\end{align}
with $H_A = H_A^{\dagger} \in \mathbb{C}^{n\times n}$, $H_k = H_k^{\dagger} \in \mathbb{C}^{k\times k}$, and $M \in \mathbb{C}^{n\times k}$. The $2nk$ real coordinates of the mixing block $M$ span the coset directions of~\eqref{eq-Omega}, and the $k^2$ coordinates of $H_k$ span the new internal block. The tower imposes exactly one constraint: $H_A$ receives zero gradient. Training touches only the shell $(M, H_k)$, and deletion projects $(M, H_k) \to 0$, which by~\eqref{eq-H-blocks} and the homomorphism property of~\eqref{eq-embed} restores every word operator, every prefix product, and the document operator to its embedded, pre-extension value.

\section{Results and discussion}\label{sec:results}

\subsection{Global vs. Adaptive Per-Word Rotation Budget}

This subsection shows how sentence sequences can be captured through the non-commutative structure of matrix multiplication, entirely passing the need for explicit positional embeddings. The core finding is that predicting an adaptive per-word phase rotation budget dynamically scales the model's expressivity and  manage longer sequences. The simulations primarily use a fixed global rotation budget constraint of $\varepsilon = 2.2$, but introducing a dynamically predicted $\varepsilon_w$ via a learned weight vector effectively allocates a tailored phase budget based on word importance. 

Table~\ref{tab:table1} compares a single global rotation budget with the
predicted per-word rotation budget of Eq.~\eqref{eq-adaptive-eps} on IMDB \cite{maas2011learning} and AG~News \cite{zhang2015character} (10{,}000-word vocabulary, 256-token cap,
three-seed means over seeds $\{1337, 42, 7\}$). Employing the predicted $\varepsilon_w$ mechanism pushes the accuracy to 86.53\% on the IMDB dataset and 87.68\% on AG News. This clearly surpasses the flat global budget baseline (85.28\% and 87.45\%, respectively) and tightly competes with the hashed bag-of-words (BoW) \cite{weinberger2009feature} baseline (85.25\% and 87.45\%). The adaptive per word rotation budget prevents phase wrap-around (aliasing) in extended sequences by appropriately down-weighting low-information tokens.

\begin{table}[h!]
\caption{\label{tab:table1}%
Accuracy comparison of adaptive per-word rotation budget arms on the IMDB and AG News datasets.
}
\begin{ruledtabular}
\begin{tabular}{lcc}
\textrm{Description}&
\textrm{IMDB}&
\textrm{AG News}\\
\colrule
Global $\varepsilon = 2.2$ & 85.28\% & 87.45\% \\
Predicted $\varepsilon_w $ as in Eq.\eqref{eq-adaptive-eps} & 86.53\% & 87.68\% \\
BoW baseline & 85.25\% & 87.45\% \\
\end{tabular}
\end{ruledtabular}
\end{table}

\subsection{Attention variants}
Parameter-efficient attention scores can be computed from prefix product operators and hermitian operator A, eliminating the need for standard query, key, and value (QKV) projections. 

The main finding is that a single learned Hermitian matrix A can generate an exactly antisymmetric attention score with robust directional awareness. The architecture harnesses an $8 \times 8$ Hermitian matrix mapping to $n^2 = 64$ parameters, which substantially reduces the conventional $3d_{\text{model}}^2$ parameter overhead while inherently capturing sequence directionality. Table~\ref{tab:table2} contrasts these QKV-free self-attention mechanisms against established conventional baselines. The operator-based QKV-free self-attention model realizes an accuracy of 84.4\% on IMDB and 87.1\% on AG News. This parameter-light structure is highly competitive, statistically matching a similarly parameterized conventional transformer (84.6\% IMDB, 84.5\% AG News) and outperforming the paired PaTH-lite method (83.4\% and 84.1\%). Although the 60k-parameter bag-of-words ceiling retains a modest edge (85.3\% and 87.6\%).

\begin{table}[h!]
\caption{\label{tab:table2}%
Accuracy comparison of attention variants on the IMDB and AG News datasets.
}
\begin{ruledtabular}
\begin{tabular}{lcc}
\textrm{Model}&
\textrm{IMDB}&
\textrm{AG News}\\
\colrule
QKV-Free Self-Attention Eq.\eqref{Sij} & 84.4\% & 87.1\% \\
Matched transformer & 84.6\% & 84.5\% \\
PaTH-lite~\cite{yang2026path} & 83.4\% & 84.1\% \\
BoW baseline & 85.3\% & 87.6\% \\
\end{tabular}
\end{ruledtabular}
\end{table}

\subsection{Chunked Attention}

Grouping consecutive words into chunk-level unitary operators allows the model to scale to longer contexts, which we examine below. By evaluating time-ordered products over consecutive words in parallel, the framework mitigates the quadratic computational cost of full token-level attention while maintaining strict group closure properties. The experiment bounds the maximum sequence length to 1024 words. A wrap-safe rotation budget of $\varepsilon/c$ is dynamically allocated per chunk of size $L/c$, confining the accumulated chunk phase to match the fundamental single-word budget. Table~\ref{tab:table3} evaluates this framework on full IMDB reviews. Building wrap-safe composites at chunk sizes of $L/c=16$ (85.8\% accuracy) and $L/c=64$ (85.6\%) retains near-perfect statistical parity with the unchunked, flat $c=1$ baseline (85.7\%). Furthermore, the $c=64$ wrap-safe configuration massively reduces the relative computational cost of the score-pair to a mere $1/4096$. In contrast, naively composing 16 tokens using the full, unscaled $\epsilon$ budget rapidly induces a phase wrap collapse, plunging accuracy down to an unrecoverable chance baseline of 51.7\%.

\begin{table}[h!]
\caption{\label{tab:table3}%
Chunked attention on the IMDB dataset (full reviews, capped at 1024 words). Wrap-safe composites are built at rotation budget $\varepsilon/c$; at the full $\varepsilon$ the composite aliases past $2\pi$ and collapses to chance.
}
\begin{ruledtabular}
\begin{tabular}{lcc}
\textrm{Configuration}&
\textrm{Mean accuracy}&
\textrm{Rel. score-pair cost}\\
\colrule
$c=1$ (flat case) & 85.7\%  & 1 \\
$c=16$ (naive full $\varepsilon$) & 51.7\% & 1/256 \\
$c=16$ (wrap-safe $\varepsilon/c$) & 85.8\%  & 1/256 \\
$c=64$ (wrap-safe $\varepsilon/c$) & 85.6\%  & 1/4096 \\
BoW baseline & 84.1\%  & --- \\
\end{tabular}
\end{ruledtabular}
\end{table}

\subsection{Reading out in Canonical Coset Coordinates}

Table~\ref{tab:table4} compares the final document operator expressed in an intrinsic geometric coordinate space against the redundant flattened tensor array \eqref{eq-u_flat}. The primary insight is that mapping the document state onto the canonical coset decomposition forms nested Householder spherical shells isolating the exact, non-redundant degrees of freedom in the matrix. Operating with a fixed rotation budget of $\varepsilon = 0.15$, the model maps its classification readout into an intrinsic 64-dimensional canonical-coset coordinate tensor, avoiding the overparameterized 128-dimensional flattening of the complex matrix entries. Table~\ref{tab:table4} reveals that utilizing the pure 64-dimensional coset readout produces reasonable accuracies of 79.4\% on IMDB and 84.8\% on AG News. However, it trails the naive 128-dimensional flattened representation (82.6\% IMDB, and 84.9\% AG News). This discrepancy is entirely attributable to the spherical boundary curvature inherent to Householder balls: longer documents progressively push the unitary points toward highly curved radial boundaries where linear decision boundaries structurally fail. Short texts (AG News) organically remain in the flatter interior, whereas fully mitigating the IMDB boundary distortion necessitates applying lower $\varepsilon$ values or adaptative per word rotation budgets $\varepsilon_w$.

\begin{table}[h!]
\caption{\label{tab:table4}%
Effect of the readout representation on mean accuracy at a fixed rotation
budget $\varepsilon = 0.15$: the canonical-coset readout (64 parameters)
against the standard flattened 128-dimensional readout
}
\begin{ruledtabular}
\begin{tabular}{lcc}
\textrm{Representation}&
\textrm{IMDB}&
\textrm{AG News}\\
\colrule
Flatten-128  & 82.6\% & 84.9\% \\
Coset-64  & 79.4\%  & 84.8\%  \\
\end{tabular}
\end{ruledtabular}
\end{table}

\subsection{Coset-tower}
\begin{table*}[t]
\caption{\label{tab:table14}%
Coset-tower continual learning on a two-task sequence, $U(8) \subset U(12)$
(shell $k = 4$, $2nk + k^2 = 80$ new real coordinates by
Eq.~\eqref{eq-shell-dim}). Task A is trained first and then frozen: $H_A$
receives zero gradient thereafter, and task B is trained entirely inside the
shell $(M, H_k)$ of Eq.~\eqref{eq-H-blocks}. Deletion is the linear projection
$(M, H_k) \to 0$ of Section~\ref{sec:coset-tower}, not a retraining step. Accuracy on A is unchanged at every
stage, whereas standard finetuning of the same model on B, with all of $H$
free, costs $30.3$ pp on A and lands at chance. Trainable coords.\ counts the
real coordinates of the generator of Eq.~\eqref{eq-H-blocks} that receive
gradient at that stage, per word operator: $n^2 = 64$ for $H_A$,
$2nk + k^2 = 80$ for the shell $(M, H_k)$, and $(n+k)^2 = 144$ for all of $H$.
The tower thus trains $80$ of $144$ coordinates and never moves the $64$ that
carry task A. Dashes mark stages where the task is undefined: B has not yet
been added (stage 1), or has been deleted (stage 3). Chance is $50\%$ on A and
$25\%$ on B. Single seed.
}
\begin{ruledtabular}
\begin{tabular}{lccc}
\textrm{Stage (trainable block)}&
\textrm{Trainable coords.}&
\textrm{Task A}&
\textrm{Task B}\\
\colrule
\multicolumn{4}{l}{\emph{Coset tower}}\\
1. Train A in $U(8)$, freeze \quad ($H_A$) & 64 & 83.5\% & --- \\
2. Embed in $U(12)$, train B \quad ($M, H_k$) & 80 & 83.5\% & 69.2\% \\
3. Delete B: $(M, H_k) \to 0$ \quad (none) & 0 & 83.5\% & 28.8\% \\
\colrule
\multicolumn{4}{l}{\emph{Reference points}}\\
Standard finetuning on B \quad (all of $H$) & 144 & 53.2\% & --- \\
B alone in unconstrained $U(12)$ & 144 & --- & 86.6\% \\
\end{tabular}
\end{ruledtabular}
\end{table*}

%***************************************************

\begin{figure*}[!t]
\centering
\adjustbox{max width=0.8\linewidth}{%
\begin{tikzpicture}[font=\footnotesize, node distance=10mm,
  blk/.style={draw, minimum size=15mm, fill=blue!12},
  shell/.style={draw, minimum size=24mm, fill=orange!10},
  ar/.style={-{Stealth[length=2mm]}, thick, gray!70},
  ph/.style={font=\footnotesize\bfseries}]
% Phase 1: u(8) frozen
\begin{scope}[local bounding box=p1]
  \node[blk] (a1) {$\mathfrak{u}(8)$};
  \node[ph, above=2mm of a1] {Phase 1};
  \node[below=1mm of a1, align=center, text=blue!50!black]
    {train A (IMDB)\\ \textbf{freeze} $\Rightarrow 83.5\%$};
\end{scope}
% Phase 2: extend to u(12), train shell
\begin{scope}[shift={(4.6,0)}, local bounding box=p2]
  \node[shell] (s2) {};
  \node[blk] (a2) at (s2.center) {};
  \node at (a2.center) {\scriptsize $\mathfrak{u}(8)$};
  \node[font=\scriptsize] at ([yshift=-9.5mm]s2.center) {new shell ($80$)};
  \node[ph, above=2mm of s2] {Phase 2};
  \node[below=1mm of s2, align=center, text=orange!55!black]
    {extend $\subset\mathfrak{u}(12)$,\\ train B in shell\\
     A still $83.5\%$ \checkmark};
\end{scope}
% Phase 3: delete shell
\begin{scope}[shift={(9.7,0)}, local bounding box=p3]
  \node[shell, dashed, fill=gray!6] (s3) {};
  \node[blk] (a3) at (s3.center) {$\mathfrak{u}(8)$};
  \node[ph, above=2mm of s3] {Phase 3};
  \node[below=1mm of s3, align=center, text=black!60]
    {project shell $\to 0$:\\ B forgotten ($28.8\%$),\\
     A intact ($83.5\%$)};
\end{scope}
\draw[ar] (p1.east) -- (p2.west);
\draw[ar] (p2.east) -- (p3.west);
\end{tikzpicture}}
\caption{ Coset-Tower and Continual Learning: The
$\mathrm{U}(8)$ block trained on domain A (blue, inner ring) is frozen and
embedded in $\mathrm{U}(12)$; domain B is
learned in the new $80$-coordinate shell (orange) without changing the inner
block, such that A is preserved exactly ($83.5\%\!\to\!83.5\%$). Whereas, standard
finetuning erase it ($\to 53.2\%$). Projecting the shell back to zero
deletes B ($\to$ chance) and leaves A unchanged.}
\label{fig:tower}
\end{figure*}
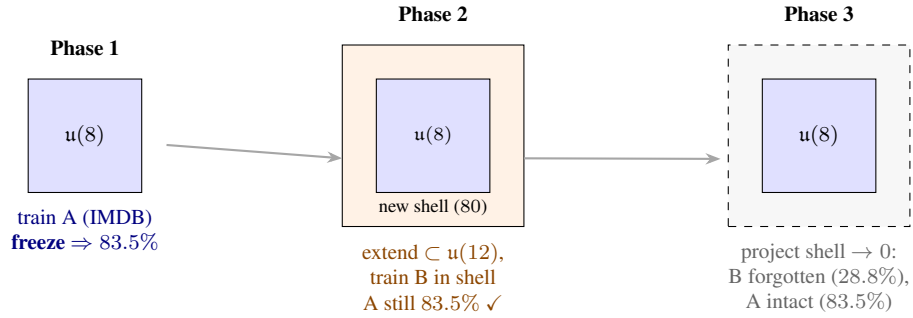

%------------------------------------------------------------------------
% Accompanying body text -- the subsection is currently empty and a PRX
% table needs prose that states the claim the numbers support.
%------------------------------------------------------------------------
Table~\ref{tab:table14} tests the coset tower on a two-task sequence. A model
trained on task A in $U(8)$ is frozen and embedded in $U(12)$ by
Eq.~\eqref{eq-embed}; task B is then trained using only the shell
coordinates $(M, H_k)$ of Eq.~\eqref{eq-H-blocks}, with $H_A$ receiving
zero gradient, as shown in Fig.(\ref{fig:tower}).

Accuracy on A is unchanged to the reported precision at
every stage. This is structural rather than empirical: the embedding is
a group homomorphism, so every word operator, every prefix product and
the document operator retain their pre-extension values exactly, with
no replay buffer and no regularization penalty. Standard finetuning of
the same model on B, with all $144$ coordinates free, drops A to
53.2\%, at chance for the binary task.

Deletion is the linear projection $(M, H_k) \to 0$ of
Section~\ref{sec:coset-readout}, reached in one step rather than by
optimization. It returns B to 28.8\%, at its 25\% chance level, while
leaving A untouched, so a task can be removed exactly and without
retraining. The cost is capacity: B reaches 69.2\% inside the
$80$-coordinate shell against 86.6\% when trained alone in the full
$U(12)$. That $17.4$ pp gap measures the part of task B that cannot be
expressed without mixing into the frozen block, and it sets the
practical limit on how far the tower can be iterated at fixed $k$.

\section{Conclusion and Outlook}\label{sec:conclusion}

In this work, we formulate a text representation framework grounded in the geometry of Lie groups. Our main contribution is to show that capabilities standard architectures achieve through separate, engineered modules arise from the algebraic properties of the underlying group action itself. The non-commutativity of matrix product encodes word
order, eliminating the need for explicit positional encodings. Building on this
structure, we derive a parameter-light, QKV-free self-attention mechanism parameterized by a single
learned Hermitian operator $A$, showing that directional sensitivity emerges directly from the
antisymmetric trace overlap of prefix states. Group closure further enables scalable, variable-length chunking that substantially reduces attention costs while preventing aliasing, making parallel composition of long texts a structural consequence. Mapping document operators into canonical-coset coordinates provides a compact, intrinsic readout that encodes all true unitary degrees of freedom,
eliminating the geometric redundancy of flattened array representations. Finally, the natural
nesting structure of unitary groups ($U_n \subset U_{n+k}$) yields a coset tower for exact continual
learning---unlike regularization approaches that rely on soft statistical constraints, this nested geometric structure preserves prior knowledge. Consequently, removing a learned task is reduced to an orthogonal projection onto a lower-dimensional subspace, avoiding a complete secondary optimization pass.

This geometric formulation opens several promising avenues for future research. While our numerical experiments focused on text classification using small operators in $\mathrm{U}(8)$, scaling the framework to higher-dimensional groups $\mathrm{U}(n)$ is be critical for increasing representational capacity. Transitioning from sequence classification to generative language modeling stands as a primary milestone. We also plan to apply the antisymmetric scoring mechanism to fundamentally directed tasks, such as question answering and natural language entailment, where asymmetrical attention is required. Additionally, integrating contextual embeddings via generator distillation could further enrich the model's semantic foundations without exploding the parameter count. Expanding the coset tower into multi-scale architectures could enable robust, hierarchical  learning systems capable of selectively auditable knowledge retention and deletion. Theoretically, our architecture inherently is able to integrate classical natural language processing with the analytical tools of random matrix theory by leveraging the statistical properties of unitary matrices. Lastly, because an $8 \times 8$ unitary matrix is mathematically equivalent to a 3-qubit quantum gate, our framework bridges classical natural language processing with quantum information theory. Although evaluated classically, this equivalence provides a seamless, first-principles blueprint for executing advanced sequence modeling on near-term quantum computing platforms.

\bibliographystyle{apsrev4-1}
\bibliography{main}

@article{su2024roformer,
  title={Roformer: Enhanced transformer with rotary position embedding},
  author={Su, Jianlin and Ahmed, Murtadha and Lu, Yu and Pan, Shengfeng and Bo, Wen and Liu, Yunfeng},
  journal={Neurocomputing},
  volume={568},
  pages={127063},
  year={2024},
  publisher={Elsevier}
}

@article{gu2023mamba,
  title={Mamba: Linear-time sequence modeling with selective state spaces},
  author={Gu, Albert and Dao, Tri},
  journal={arXiv preprint arXiv:2312.00752},
  year={2023}
}

@article{gu2021efficiently,
  title={Efficiently modeling long sequences with structured state spaces},
  author={Gu, Albert and Goel, Karan and R{\'e}, Christopher},
  journal={arXiv preprint arXiv:2111.00396},
  year={2021}
}

@article{saponati2025underlying,
  title={The underlying structures of self-attention: symmetry, directionality, and emergent dynamics in Transformer training},
  author={Saponati, Matteo and Sager, Pascal and Aceituno, Pau Vilimelis and Stadelmann, Thilo and Grewe, Benjamin},
  journal={arXiv preprint arXiv:2502.10927},
  year={2025}
}

@inproceedings{katharopoulos2020transformers,
  title={Transformers are rnns: Fast autoregressive transformers with linear attention},
  author={Katharopoulos, Angelos and Vyas, Apoorv and Pappas, Nikolaos and Fleuret, Fran{\c{c}}ois},
  booktitle={International conference on machine learning},
  pages={5156--5165},
  year={2020},
  organization={PMLR}
}

@article{kayyam2026transformers,
  title={Do Transformers Need Three Projections? Systematic Study of QKV Variants},
  author={Kayyam, Ali and Gopal, Anusha Madan and Lewis, M Anthony},
  journal={arXiv preprint arXiv:2606.04032},
  year={2026}
}

@article{yang2026path,
  title={Path attention: Position encoding via accumulating householder transformations},
  author={Yang, Songlin and Shen, Yikang and Wen, Kaiyue and Tan, Shawn and Mishra, Mayank and Ren, Liliang and Panda, Rameswar and Kim, Yoon},
  journal={Advances in Neural Information Processing Systems},
  volume={38},
  pages={62220--62247},
  year={2026}
}

@article{cabrera2010canonical,
  title={The canonical coset decomposition of unitary matrices through Householder transformations},
  author={Cabrera, Renan and Strohecker, Traci and Rabitz, Herschel},
  journal={Journal of Mathematical Physics},
  volume={51},
  number={8},
  year={2010},
  publisher={AIP Publishing}
}

@article{householder1958unitary,
  title={Unitary triangularization of a nonsymmetric matrix},
  author={Householder, Alston S},
  journal={Journal of the ACM (JACM)},
  volume={5},
  number={4},
  pages={339--342},
  year={1958},
  publisher={ACM New York, NY, USA}
}

@inproceedings{arjovsky2016unitary,
  title={Unitary evolution recurrent neural networks},
  author={Arjovsky, Martin and Shah, Amar and Bengio, Yoshua},
  booktitle={International conference on machine learning},
  pages={1120--1128},
  year={2016},
  organization={PMLR}
}

@inproceedings{jing2017tunable,
  title={Tunable efficient unitary neural networks (EUNN) and their application to RNNs},
  author={Jing, Li and Shen, Yichen and Dubcek, Tena and Peurifoy, John and Skirlo, Scott and LeCun, Yann and Tegmark, Max and Solja{\v{c}}i{\'c}, Marin},
  booktitle={International Conference on Machine Learning},
  pages={1733--1741},
  year={2017},
  organization={PMLR}
}

@article{kerg2019non,
  title={Non-normal recurrent neural network (nnrnn): learning long time dependencies while improving expressivity with transient dynamics},
  author={Kerg, Giancarlo and Goyette, Kyle and Puelma Touzel, Maximilian and Gidel, Gauthier and Vorontsov, Eugene and Bengio, Yoshua and Lajoie, Guillaume},
  journal={Advances in neural information processing systems},
  volume={32},
  year={2019}
}

@article{bernardy2022unitary,
  title={Unitary recurrent networks: Algebraic and linear structures for syntax},
  author={Bernardy, Jean-Philippe and Lappin, Shalom},
  journal={Algebraic Structures in Natural Language},
  pages={243--278},
  year={2022},
  publisher={CRC Press}
}

@article{bernardy2022assessing,
  title={Assessing the unitary rnn as an end-to-end compositional model of syntax},
  author={Bernardy, Jean-Philippe and Lappin, Shalom},
  journal={arXiv preprint arXiv:2208.05719},
  year={2022}
}

@article{fan2024quantum,
  title={Quantum-inspired language models based on unitary transformation},
  author={Fan, Zipeng and Zhang, Jing and Zhang, Peng and Lin, Qianxi and Li, Yizhe and Qian, Yuhua},
  journal={Information Processing \& Management},
  volume={61},
  number={4},
  pages={103741},
  year={2024},
  publisher={Elsevier}
}

@article{anschuetz2023interpretable,
  title={Interpretable quantum advantage in neural sequence learning},
  author={Anschuetz, Eric R and Hu, Hong-Ye and Huang, Jin-Long and Gao, Xun},
  journal={PRX Quantum},
  volume={4},
  number={2},
  pages={020338},
  year={2023},
  publisher={APS}
}

@article{nausheen2025quantum,
  title={Quantum natural language processing: A comprehensive review of models, methods, and applications},
  author={Nausheen, Farha and Ahmed, Khandakar and Khan, M Imad and Riaz, Farina},
  journal={arXiv preprint arXiv:2504.09909},
  year={2025}
}

@misc{gilmore2011group,
  title={Group Theory: A Physicist’s Survey},
  author={Gilmore, Robert},
  year={2011},
  publisher={American Institute of Physics}
}

@book{cahn2014semi,
  title={Semi-simple Lie algebras and their representations},
  author={Cahn, Robert N},
  year={2014},
  publisher={Courier Corporation}
}

@inproceedings{zhang2024learning,
  title={Learning disentangled semantic spaces of explanations via invertible neural networks},
  author={Zhang, Yingji and Carvalho, Danilo and Freitas, Andre},
  booktitle={Proceedings of the 62nd Annual Meeting of the Association for Computational Linguistics (Volume 1: Long Papers)},
  pages={2113--2134},
  year={2024}
}

@article{vaswani2017attention,
  title={Attention is all you need},
  author={Vaswani, Ashish and Shazeer, Noam and Parmar, Niki and Uszkoreit, Jakob and Jones, Llion and Gomez, Aidan N and Kaiser, {\L}ukasz and Polosukhin, Illia},
  journal={Advances in neural information processing systems},
  volume={30},
  year={2017}
}

@inproceedings{maas2011learning,
  title={Learning word vectors for sentiment analysis},
  author={Maas, Andrew and Daly, Raymond E and Pham, Peter T and Huang, Dan and Ng, Andrew Y and Potts, Christopher},
  booktitle={Proceedings of the 49th annual meeting of the association for computational linguistics: Human language technologies},
  pages={142--150},
  year={2011}
}

@article{zhang2015character,
  title={Character-level convolutional networks for text classification},
  author={Zhang, Xiang and Zhao, Junbo and LeCun, Yann},
  journal={Advances in neural information processing systems},
  volume={28},
  year={2015}
}

@article{mikolov2013efficient,
  title={Efficient estimation of word representations in vector space},
  author={Mikolov, Tomas and Chen, Kai and Corrado, Greg and Dean, Jeffrey},
  journal={arXiv preprint arXiv:1301.3781},
  year={2013}
}

@article{kirkpatrick2017overcoming,
  title={Overcoming catastrophic forgetting in neural networks},
  author={Kirkpatrick, James and Pascanu, Razvan and Rabinowitz, Neil and Veness, Joel and Desjardins, Guillaume and Rusu, Andrei A and Milan, Kieran and Quan, John and Ramalho, Tiago and Grabska-Barwinska, Agnieszka and others},
  journal={Proceedings of the national academy of sciences},
  volume={114},
  number={13},
  pages={3521--3526},
  year={2017},
  publisher={National Academy of Sciences}
}

@article{lopez2017gradient,
  title={Gradient episodic memory for continual learning},
  author={Lopez-Paz, David and Ranzato, Marc'Aurelio},
  journal={Advances in neural information processing systems},
  volume={30},
  year={2017}
}

@article{rusu2016progressive,
  title={Progressive neural networks},
  author={Rusu, Andrei A and Rabinowitz, Neil C and Desjardins, Guillaume and Soyer, Hubert and Kirkpatrick, James and Kavukcuoglu, Koray and Pascanu, Razvan and Hadsell, Raia},
  journal={arXiv preprint arXiv:1606.04671},
  year={2016}
}

@inproceedings{mallya2018packnet,
  title={Packnet: Adding multiple tasks to a single network by iterative pruning},
  author={Mallya, Arun and Lazebnik, Svetlana},
  booktitle={2018 IEEE/CVF Conference on Computer Vision and Pattern Recognition},
  pages={7765--7773},
  year={2018},
  organization={IEEE}
}

@inproceedings{serra2018overcoming,
  title={Overcoming catastrophic forgetting with hard attention to the task},
  author={Serra, Joan and Suris, Didac and Miron, Marius and Karatzoglou, Alexandros},
  booktitle={International conference on machine learning},
  pages={4548--4557},
  year={2018},
  organization={PMLR}
}

@inproceedings{ioffe2015batch,
  title={Batch normalization: Accelerating deep network training by reducing internal covariate shift},
  author={Ioffe, Sergey and Szegedy, Christian},
  booktitle={International conference on machine learning},
  pages={448--456},
  year={2015},
  organization={pmlr}
}

@inproceedings{weinberger2009feature,
  title={Feature hashing for large scale multitask learning},
  author={Weinberger, Kilian and Dasgupta, Anirban and Langford, John and Smola, Alex and Attenberg, Josh},
  booktitle={Proceedings of the 26th annual international conference on machine learning},
  pages={1113--1120},
  year={2009}
}

@inproceedings{pascanu2013difficulty,
  title={On the difficulty of training recurrent neural networks},
  author={Pascanu, Razvan and Mikolov, Tomas and Bengio, Yoshua},
  booktitle={International conference on machine learning},
  pages={1310--1318},
  year={2013},
  organization={Pmlr}
}

@book{nielsen2001quantum,
  title={Quantum computation and quantum information},
  author={Nielsen, Michael A and Chuang, Isaac L},
  volume={2},
  year={2001},
  publisher={Cambridge university press Cambridge}
}

\appendix \onecolumngrid
Appendix~\ref{app:bounding-eigenphase}

\section{Bounding $\varepsilon_w$} \label{app:bounding-eigenphase}

This appendix derives the eigenphase bound $|\lambda_a| \leq \varepsilon_w$ quoted in the main text, which follows from the normalization of the Hermitian generator to unit Frobenius norm, and then makes precise how the per-word eigenphases accumulate under the prefix product of Eq.~\ref{eq-prefix}, quantifying the wrap-safety condition invoked by the chunked construction of Eq.~\ref{eq-chunk}. The bound ensures that each word induces a small, controlled rotation on the unitary manifold, preventing phase wrap and maintaining numerical stability.
 
\emph{The single-word bound}: Let $H_w \in \mathbb{C}^{n \times n}$, $H_w = H_w^\dagger$, be the learned Hermitian generator of word $w$, with Frobenius norm
\begin{align}
\lVert H_w \rVert_F \;=\; \sqrt{\operatorname{Tr}\!\big(H_w^\dagger H_w\big)} \;=\; \Big(\sum_{i,j} \big|(H_w)_{ij}\big|^2\Big)^{\!1/2},
\end{align}
and let $\hat{H}_w = H_w / \lVert H_w \rVert_F$ denote its normalization to unit Frobenius norm, $\lVert \hat{H}_w \rVert_F = 1$, as used in Eq.~\eqref{eq-Uw}. By the spectral theorem, $\hat{H}_w$ has real eigenvalues $\hat{\lambda}_1, \ldots, \hat{\lambda}_n$, and since $\hat{H}_w^\dagger \hat{H}_w = \hat{H}_w^2$, the Frobenius norm of a Hermitian matrix equals the Euclidean norm of its eigenvalue vector,
\begin{align}
\lVert \hat{H}_w \rVert_F^2 \;=\; \operatorname{Tr}\big(\hat{H}_w^2\big) \;=\; \sum_{a=1}^{n} \hat{\lambda}_a^2 \;=\; 1 .
\end{align}
The effective generator entering Eq.~\eqref{eq-Uw} is $\varepsilon_w \hat{H}_w$; its eigenvalues $\lambda_a = \varepsilon_w \hat{\lambda}_a$ therefore obey the sum rule
\begin{align}
\label{eq-sum-rule}
\sum_{a=1}^{n} \lambda_a^2 \;=\; \varepsilon_w^2 .
\end{align}
Because every term in Eq.~\eqref{eq-sum-rule} is non-negative, each term is bounded by the full sum,
\begin{align}
\label{eq-eigenphase-bound}
\lambda_a^2 \;\leq\; \sum_{b=1}^{n} \lambda_b^2 \;=\; \varepsilon_w^2
\qquad \Longrightarrow \qquad
|\lambda_a| \;\leq\; \varepsilon_w , \qquad a = 1, \ldots, n .
\end{align}
Equivalently, $\max_a |\lambda_a| = \lVert \varepsilon_w \hat{H}_w \rVert_2 \leq \lVert \varepsilon_w \hat{H}_w \rVert_F = \varepsilon_w$, where $\lVert \cdot \rVert_2$ denotes the operator (spectral) norm, which for Hermitian matrices coincides with the spectral radius. The bound is saturated, $|\lambda_a| = \varepsilon_w$, if and only if all remaining eigenvalues vanish, i.e., if and only if $\hat{H}_w = \pm \lvert v \rangle\langle v \rvert$ is rank one for some unit vector $\lvert v \rangle$; for generic full-rank generators the normalization spreads the budget over the whole spectrum, and every individual eigenphase is strictly smaller than $\varepsilon_w$.
 
Since $U_w = \exp(i \varepsilon_w \hat{H}_w)$ is diagonal in the eigenbasis of $\hat{H}_w$, its eigenvalues are $e^{i \lambda_a}$: the eigenphases of the word operator are exactly the $\lambda_a$, confined by Eq.~\eqref{eq-eigenphase-bound} to the interval $[-\varepsilon_w, \varepsilon_w]$. Provided $\varepsilon_w < \pi$, this interval lies strictly inside the principal branch $(-\pi, \pi]$, so a single word operator never aliases its phases: each word induces a small, unambiguous rotation whose magnitude is set by the budget $\varepsilon_w$ alone.
 
\emph{Phase accumulation over a document}: Consider a sequence $\{w_1, \ldots, w_L\}$ with per-word budgets $\varepsilon_{w_i}$ given by Eq.~\eqref{eq-adaptive-eps}, word operators $U_i = \exp(i \varepsilon_{w_i} \hat{H}_{w_i})$, and document operator $P_L = U_L \cdots U_1$ as in Eq.~\eqref{eq-prefix}. Write $\lambda_a^{(i)} = \varepsilon_{w_i} \hat{\lambda}_a^{(i)}$ for the eigenvalues of the $i$-th effective generator $\varepsilon_{w_i} \hat{H}_{w_i}$, and denote the total rotation budget by
\begin{align}
\Phi_L \;\equiv\; \sum_{i=1}^{L} \varepsilon_{w_i} .
\end{align}
If all generators commuted, then $P_L = \exp\big( i \sum_{i=1}^{L} \varepsilon_{w_i} \hat{H}_{w_i} \big)$ would hold exactly and the eigenphases would be literally additive, $\theta_a = \sum_{i=1}^{L} \lambda_a^{(i)} \pmod{2\pi}$, which is the linear accumulation $\sum_{i=1}^{L} \varepsilon_{w_i} \hat{\lambda}_a^{(i)}$ quoted in the main text; the single-word bound~\eqref{eq-eigenphase-bound} and the triangle inequality then give $\big|\sum_i \lambda_a^{(i)}\big| \leq \Phi_L$. For non-commuting generators the eigenphases of a product are no longer sums of the eigenphases of its factors: the departure from additivity, entering at order $\varepsilon_{w_i} \varepsilon_{w_j} \big[ \hat{H}_{w_i}, \hat{H}_{w_j} \big]$ in the Baker--Campbell--Hausdorff expansion of $\log P_L$, is precisely the structure that encodes word order. The bound, however, survives non-commutativity. Telescoping the product,
\begin{align}
P_L - I \;=\; \sum_{k=1}^{L} U_L \cdots U_{k+1} \big( U_k - I \big),
\end{align}
and using the unitary invariance of the operator norm together with the single-word spectral bound,
$\lVert U_k - I \rVert_2 = \max_a \big| e^{i \lambda_a^{(k)}} - 1 \big| = 2 \max_a \big| \sin\big(\lambda_a^{(k)}/2\big) \big| \leq \max_a \big|\lambda_a^{(k)}\big| \leq \varepsilon_{w_k}$, yields
\begin{align}
\label{eq-product-bound}
\lVert P_L - I \rVert_2 \;\leq\; \sum_{k=1}^{L} \lVert U_k - I \rVert_2 \;\leq\; \sum_{k=1}^{L} \varepsilon_{w_k} \;=\; \Phi_L .
\end{align}
Since $P_L \in \mathrm{U}(n)$, its eigenvalues are $e^{i \theta_a}$ with principal eigenphases $\theta_a \in (-\pi, \pi]$, and $P_L - I$ is normal, so $\max_a |e^{i \theta_a} - 1| = \lVert P_L - I \rVert_2$. Combining this with Eq.~\eqref{eq-product-bound} gives $2\,|\sin(\theta_a/2)| \leq \Phi_L$, and hence, whenever $\Phi_L < 2$,
\begin{align}
\label{eq-wrap-bound}
|\theta_a| \;\leq\; 2 \arcsin\!\big( \Phi_L / 2 \big) \;\leq\; \frac{\pi}{2}\, \Phi_L ,
\qquad
2 \arcsin\!\big( \Phi_L / 2 \big) \;=\; \Phi_L + O\big( \Phi_L^3 \big) .
\end{align}
In particular, $\Phi_L < 2$ guarantees $|\theta_a| < \pi$ for every eigenphase of the document operator: the spectrum of $P_L$ remains strictly inside the principal branch, no $2\pi$ wrap can occur, and for small total budget the accumulated eigenphase is bounded by $\Phi_L$ itself up to cubic corrections. The linear bound~\eqref{eq-product-bound} is a worst case, attained only when successive words rotate a common eigenvector coherently in the same direction; for generic non-commuting generators the increments partially cancel, and the accumulated phase grows more slowly than $\Phi_L$.
 
\emph{Wrap-safe chunking and the adaptive budget}: With a uniform budget, $\varepsilon_{w_i} = \varepsilon$, the total is $\Phi_L = L \varepsilon$ and grows linearly with document length, which is why a fixed global $\varepsilon$ drives long documents toward wrap-around. The chunking rule of Eq.~\eqref{eq-chunk}, $\varepsilon_w = \varepsilon / c$, caps the accumulated budget of each ordered product at $c \cdot (\varepsilon / c) = \varepsilon$ independently of the chunk size $c$, so by Eq.~\eqref{eq-wrap-bound} every chunk operator $\widetilde{P}_k$ is wrap-safe provided $\varepsilon < 2$; this makes quantitative the requirement that $\varepsilon$ sit below the wrap threshold. The adaptive mechanism of Eq.~\eqref{eq-adaptive-eps} instead replaces the uniform total $L \varepsilon$ by the learned quantity $\Phi_L = \sum_{i=1}^{L} \varepsilon_{w_i}$, allowing the network to assign small increments to low-information function words and to spend the phase budget on discriminative content words, so that the accumulated phase tracks the information content of a document rather than its raw length. All of the bounds above hold verbatim with these word-dependent budgets.

\section{Spectral and density-operator interpretation of $A$.} \label{specA}

A single learned operator $A$, contracted as $\mathrm{Tr}(A\,P_i^\dagger P_j)$, selects which aspects of the relative rotation contribute to the attention score. Writing the spectral decomposition $A = \sum_k \lambda_k \lvert k \rangle\langle k \rvert$, with real eigenvalues $\lambda_k$ and orthonormal eigenvectors $\lvert k \rangle$, makes the role of $A$ concrete
\begin{align}
\label{eq-spectral-A}
\mathrm{Tr}\!\big(A\, P_i^\dagger P_j\big) \;=\; \sum_k \lambda_k\, \langle P_i k \vert P_j k \rangle.
\end{align}
This is a weighted sum of overlaps between the reference states $\lvert k \rangle$ transported by $P_i$ and by $P_j$. The constraint $A^\dagger = A$ is necessary rather than optional, as Hermiticity is the precise condition under which the imaginary part of the contraction remains exactly antisymmetric.  

This score admits a natural interpretation within the quantum-mechanical framework \cite{nielsen2001quantum}: the expectation value $\mathrm{Tr}(\rho\, M)$ extracts a scalar from an operator $M$ given a state $\rho$. When $A$ additionally satisfies positive semidefiniteness and unit trace, cycling the trace through $\rho = \rho^{1/2}\rho^{1/2}$ yields $\mathrm{Tr}(\rho\, P_i^\dagger P_j) = \mathrm{Tr}\big((P_i \rho^{1/2})^\dagger P_j \rho^{1/2}\big)$. The learned $A$ is accordingly a generalization of a mixed-state density operator with positivity and trace normalization relaxed, retaining only the structurally required Hermiticity $A^\dagger = A$. Enforcing $A = BB^\dagger$ with $\mathrm{Tr}(BB^\dagger) = 1$ would make the
identification with a learned mixed state exact, at the cost of constraining
the score to a positive spectrum; we leave $A$ unconstrained beyond
Hermiticity so that the eigenvalues $\lambda_k$ of Eq.~(\ref{eq-spectral-A})
may take either sign, allowing the score to subtract as well as add overlap
contributions.

\section{Householder Construction of the Coset Coordinates and the Ball Constraint} \label{app:coset-construction}
Theorem~1 of~\cite{cabrera2010canonical} expresses each factor of Eq.~(\ref{eq-chain}) as a product of two Householder reflections~\cite{householder1958unitary},
\begin{align}
\frac{U(n-k+1)}{U(n-k)\otimes U(1)} \;=\; R_{\lvert u_k \rangle}\, R_{\lvert e_k \rangle},
\qquad
R_{\lvert u \rangle} \;=\; I - 2\,\frac{\lvert u \rangle\langle u \rvert}{\langle u \vert u \rangle},
\end{align}
where $R_{\lvert e_k \rangle}$ is the identity with its $k$-th diagonal entry replaced by $-1$, and the Householder vector
\begin{align}
\lvert u_k \rangle \;=\; \lvert W_k \rangle + e^{i\varphi_k}\lvert e_k \rangle,
\qquad
\varphi_k \;=\; \arg\langle e_k \vert W_k \rangle,
\end{align}
is built from the $k$-th column $\lvert W_k \rangle$ of the partially reduced operator $R_{\lvert u_{k-1} \rangle}\cdots R_{\lvert u_1 \rangle} P_L$. Expanding the product $R_{\lvert u_k \rangle} R_{\lvert e_k \rangle}$ and collecting blocks yields exactly the matrix of Eq.~(\ref{eq-U_cotient})~\cite{cabrera2010canonical}.

\emph{The ball constraint}: The constraint $\langle X_k \vert X_k \rangle \leq 1$ follows from the unit length of the columns. Split the current column into its component along the target axis and the perpendicular remainder,
\begin{align}
\lvert W_k \rangle \;=\; \rho_k\, e^{i\varphi_k}\lvert e_k \rangle \;+\; \lvert W_{k\perp} \rangle,
\qquad \rho_k \geq 0.
\end{align}
Normalizing the Householder vector gives $\lvert n_k \rangle = \lvert u_k \rangle / \sqrt{\langle u_k \vert u_k \rangle} = \gamma_k \lvert e_k \rangle + \lvert n_{k\perp} \rangle$, with $\langle u_k \vert u_k \rangle = 2(1 + \rho_k)$ and $\gamma_k = \sqrt{(1+\rho_k)/2}\; e^{i\varphi_k}$. The coset vector appearing in Eq.~(\ref{eq-U_cotient}) is $\lvert X_k \rangle = 2\gamma_k^{*} \lvert n_{k\perp} \rangle$~\cite{cabrera2010canonical}, which simplifies to
\begin{align}
\lvert X_k \rangle \;=\; e^{-i\varphi_k}\, \lvert W_{k\perp} \rangle :
\end{align}
the perpendicular part of the column with its leading phase stripped off. Since $\lvert W_k \rangle$ has unit norm, $\rho_k^2 + \langle W_{k\perp} \vert W_{k\perp} \rangle = 1$, and therefore
\begin{align}
\langle X_k \vert X_k \rangle \;=\; 1 - \rho_k^2 \;\leq\; 1.
\end{align}
In real Cartesian coordinates $(x_k^1, \ldots, x_k^{2(n-k)})$ the coset vector consequently ranges over the closed unit ball
\begin{align}
B^{d} \;=\; \Bigl\{\, x \in \mathbb{R}^{d} \;:\; \textstyle\sum_{j=1}^{d} (x^j)^2 \leq 1 \,\Bigr\},
\qquad d = 2(n-k),
\end{align}
and $\sqrt{1 - r_k^2} = \rho_k$ reproduces the top-left entry of Eq.~(\ref{eq-U_cotient}): the radius measures how far the column tilts away from its axis, and what remains of the square root is the surviving alignment.

\emph{The sweep and its inverse:} The full chart is produced by one Householder sweep: reflecting with $R_{\lvert u_k \rangle}$ aligns the $k$-th column, the recursion continues on the remaining $(n-k)\times(n-k)$ block, and at each step $\lvert X_k \rangle$ and $\varphi_k$ are read off, with the $n$ residual phases collected in the diagonal factor of Eq.~(\ref{eq-chain}). The inverse map is equally explicit (Corollary~2 of~\cite{cabrera2010canonical}): the normalized Householder vector is reconstructed from the coset vector as
\begin{align}
\lvert n_k \rangle \;=\; \gamma_k \lvert e_k \rangle + \frac{1}{2\gamma_k^{*}}\lvert X_k \rangle,
\qquad
\lvert \gamma_k \rvert^2 \;=\; \frac{1 + \sqrt{1 - \langle X_k \vert X_k \rangle}}{2},
\end{align}
with the phase of $\gamma_k$ taken from the diagonal factor. This explicit inverse is what makes the round-trip reconstruction of Section~\ref{sec:coset-readout} exact to numerical precision ($1.1 \times 10^{-15}$).

\end{document}